\documentclass[pdflatex,sn-mathphys-num]{sn-jnl}%
\usepackage{graphicx}%
\usepackage{multirow}%
\usepackage{amsmath,amssymb,amsfonts}%
\usepackage{amsthm}%
\usepackage{mathrsfs}%
\usepackage[title]{appendix}%
\usepackage{xcolor}%
\usepackage{textcomp}%
\usepackage{manyfoot}%
\usepackage{booktabs}%
\usepackage{graphicx}   

\usepackage{algorithmicx}%
\usepackage{algpseudocode}%
\usepackage{listings}%
\usepackage[ruled,vlined]{algorithm2e}
\usepackage{listings}%

\usepackage{subfig}
\usepackage{tikz}
\usetikzlibrary{shapes.geometric, arrows}

\tikzstyle{startstop} = [rectangle, rounded corners, minimum width=3cm, minimum height=1cm,text centered, draw=black, fill=red!30]
\tikzstyle{process} = [rectangle, minimum width=3cm, minimum height=1cm, text centered, draw=black, fill=blue!20]
\tikzstyle{decision} = [diamond, minimum width=3cm, minimum height=1cm, text centered, draw=black, fill=green!30]
\tikzstyle{arrow} = [thick,->,>=stealth]

\begin{document}

\title{Four-Stage Alzheimer’s Disease Classification from MRI Using Topological Feature Extraction, Feature Selection, and Ensemble Learning}

\author*[1]{\fnm{Faisal Ahmed} }\email{ahmedf9@erau.edu}

\affil*[1]{\orgdiv{Department of Data Science and Mathematics}, \orgname{Embry-Riddle Aeronautical University}, \orgaddress{\street{3700 Willow Creek Rd}, \city{Prescott}, \postcode{86301}, \state{Arizona}, \country{USA}}}

\abstract{

Accurate and efficient classification of Alzheimer’s disease (AD) severity from brain magnetic resonance imaging (MRI) remains a critical challenge, particularly when limited data and model interpretability are of concern. In this work, we propose TDA-Alz, a novel framework for four-stage Alzheimer’s disease severity classification (non-demented, moderate dementia, mild, and very mild) using topological data analysis (TDA) and ensemble learning. Instead of relying on deep convolutional architectures or extensive data augmentation, our approach extracts topological descriptors that capture intrinsic structural patterns of brain MRI, followed by feature selection to retain the most discriminative topological features. These features are then classified using an ensemble learning strategy to achieve robust multiclass discrimination.

Experiments conducted on the OASIS-1 MRI dataset demonstrate that the proposed method achieves an accuracy of 98.19\% and an AUC of 99.75\%, outperforming or matching state-of-the-art deep learning–based methods reported on OASIS and OASIS-derived datasets. Notably, the proposed framework does not require data augmentation, pretrained networks, or large-scale computational resources, making it computationally efficient and fast compared to deep neural network approaches. Furthermore, the use of topological descriptors provides greater interpretability, as the extracted features are directly linked to the underlying structural characteristics of brain MRI rather than opaque latent representations. These results indicate that TDA-Alz offers a powerful, lightweight, and interpretable alternative to deep learning models for MRI-based Alzheimer’s disease severity classification, with strong potential for real-world clinical decision-support systems.

}

\keywords{Alzheimer’s disease, Magnetic resonance imaging (MRI), Topological data analysis, Ensemble learning, Disease severity classification}



\maketitle

\section{Introduction}\label{sec1}
Alzheimer’s disease (AD) is a progressive neurodegenerative disorder and one of the leading causes of dementia worldwide, characterized by gradual cognitive decline and structural brain deterioration. Early and accurate identification of AD severity stages is crucial for timely clinical intervention and disease management. Magnetic Resonance Imaging (MRI) is widely employed as a non-invasive neuroimaging modality to analyze brain morphology and structural alterations associated with Alzheimer’s disease, particularly changes in cortical thickness, hippocampal volume, and ventricular enlargement~\cite{jack2008alzheimer, mosconi2005early}.

Despite the clinical relevance of MRI, automated analysis of brain MRI for Alzheimer’s disease classification remains challenging. These challenges stem from the subtle and progressive nature of structural changes across disease stages, high inter-subject variability, and the limited availability of well-annotated medical imaging datasets~\cite{liu2021deep}. Conventional machine learning methods often rely on handcrafted features that may fail to capture complex structural relationships, while deep learning approaches typically require large-scale annotated data to achieve reliable performance.

In recent years, convolutional neural networks (CNNs) and deep learning models have demonstrated strong performance in MRI-based Alzheimer’s disease classification~\cite{zhang2021multi}. However, these models primarily focus on local intensity-based features and often lack robustness when trained on small or imbalanced datasets. Furthermore, deep models are computationally expensive, heavily dependent on data augmentation strategies, and are frequently criticized for their limited interpretability, which restricts their adoption in clinical settings~\cite{hernandez2019ensemble}.

Topological Data Analysis (TDA) has emerged as a powerful mathematical framework for extracting global and shape-aware features from complex data~\cite{carlsson2009topology}. By leveraging concepts such as persistent homology, TDA captures intrinsic structural patterns that remain stable under noise and small perturbations~\cite{edelsbrunner2008persistent}. In the context of medical imaging, TDA has shown promise in characterizing anatomical structures and disease-related patterns in a robust and interpretable manner~\cite{ahmed2023topo}. Unlike deep learning approaches, topological descriptors summarize global geometric properties of brain structures and do not require large training datasets.

Motivated by these advantages, we propose \textbf{TDA-Alz}, a fast and interpretable framework for \textbf{four-stage Alzheimer’s disease severity classification} from brain MRI. The proposed method extracts discriminative \textbf{topological descriptors} from MRI images using persistent homology, followed by \textbf{feature selection} to retain the most informative topological features. These selected features are then classified using an \textbf{ensemble learning} strategy to improve robustness and generalization. Importantly, the proposed framework does not rely on data augmentation, pretrained networks, or computationally intensive training procedures.

\begin{figure*}[t!]
    \centering
    \subfloat[\scriptsize 3D PCA projection of Betti-0 features showing clear separation among the four Alzheimer’s disease classes.\label{fig:pca-B0}]{
        \includegraphics[width=0.45\linewidth]{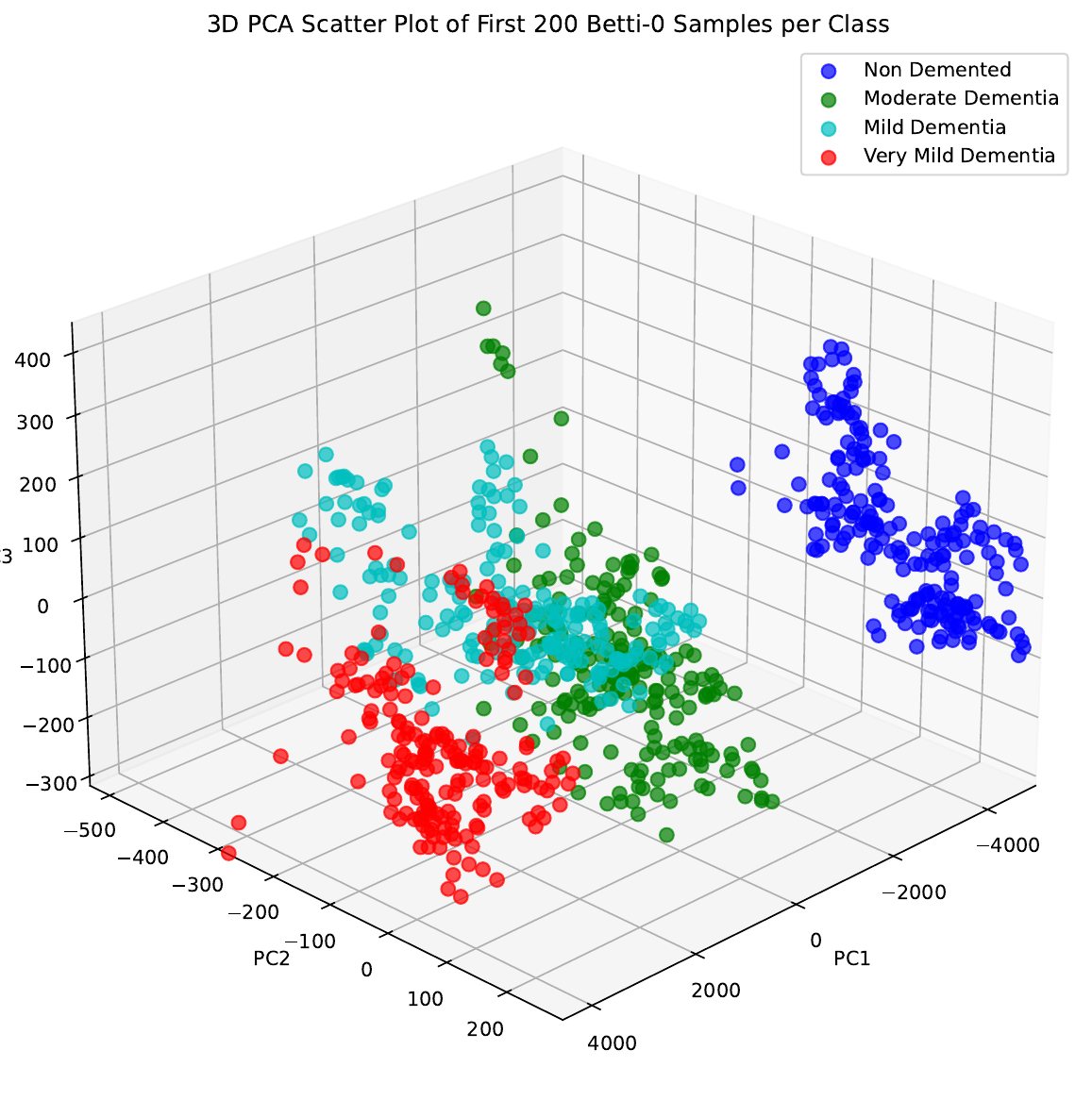}}
    \hfill
    \subfloat[\scriptsize 3D PCA projection of Betti-1 features illustrating distinct clustering of each Alzheimer’s disease class.\label{fig:pca-B1}]{
        \includegraphics[width=0.45\linewidth]{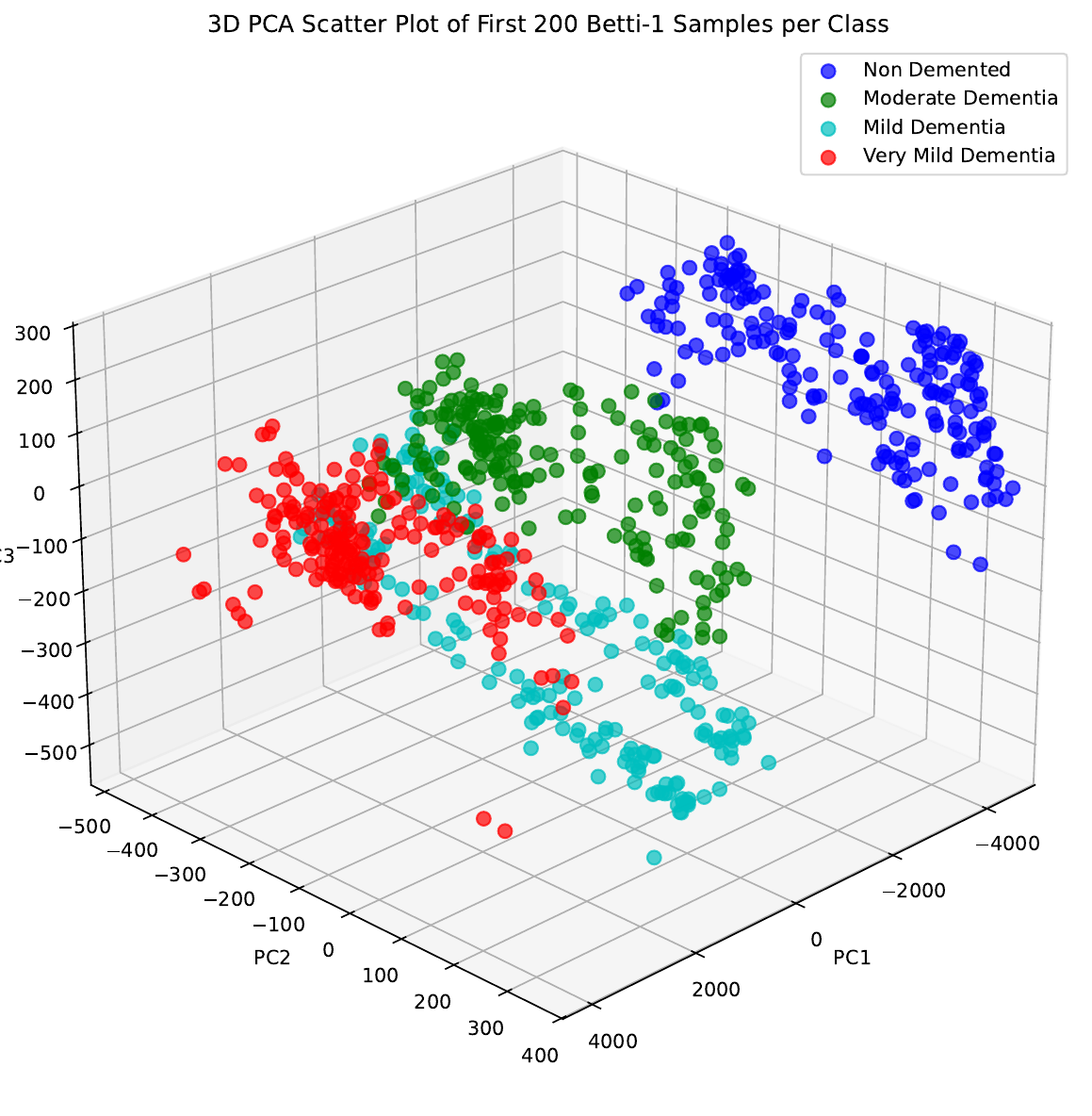}}
    
    \caption{\footnotesize Visual analysis of topological features using the first three principal components. (a) Betti-0 features and (b) Betti-1 features demonstrate clear and distinct clustering corresponding to the four AD severity stages, highlighting the discriminative power of topological descriptors.}
    \label{fig:pca-visualization}
\end{figure*}

\begin{figure*}[t!]
    \centering
    \subfloat[\scriptsize Violin plot of Betti-0 (connected components) features across the four Alzheimer’s disease stages, revealing clearly separated and class-specific distributional patterns.\label{fig:vio-B0}]{
        \includegraphics[width=0.45\linewidth]{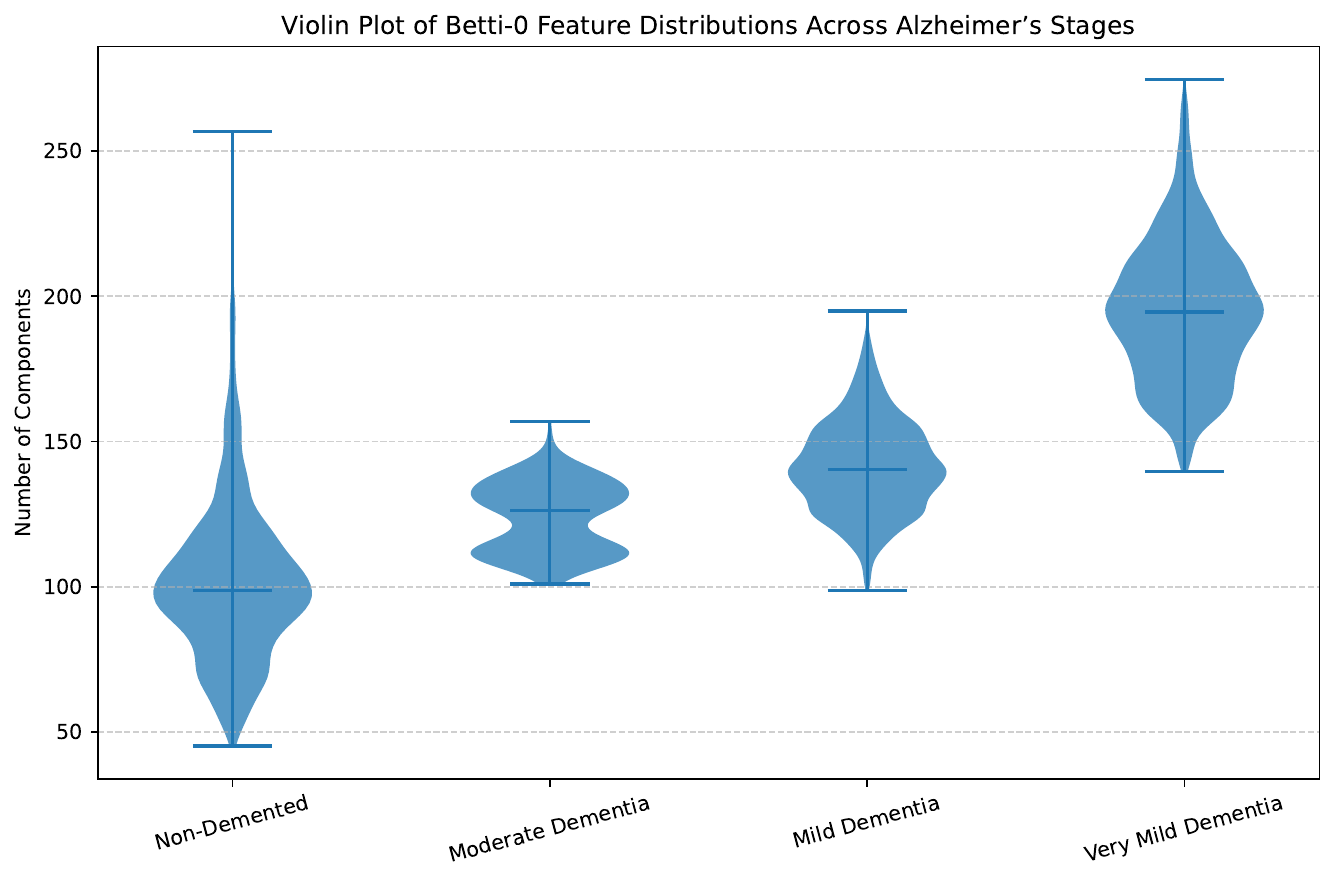}}
    \hfill
    \subfloat[\scriptsize Violin plot of Betti-1 (topological loops) features showing distinct, well-structured distributions for each Alzheimer’s disease class.\label{fig:vio-B1}]{
        \includegraphics[width=0.45\linewidth]{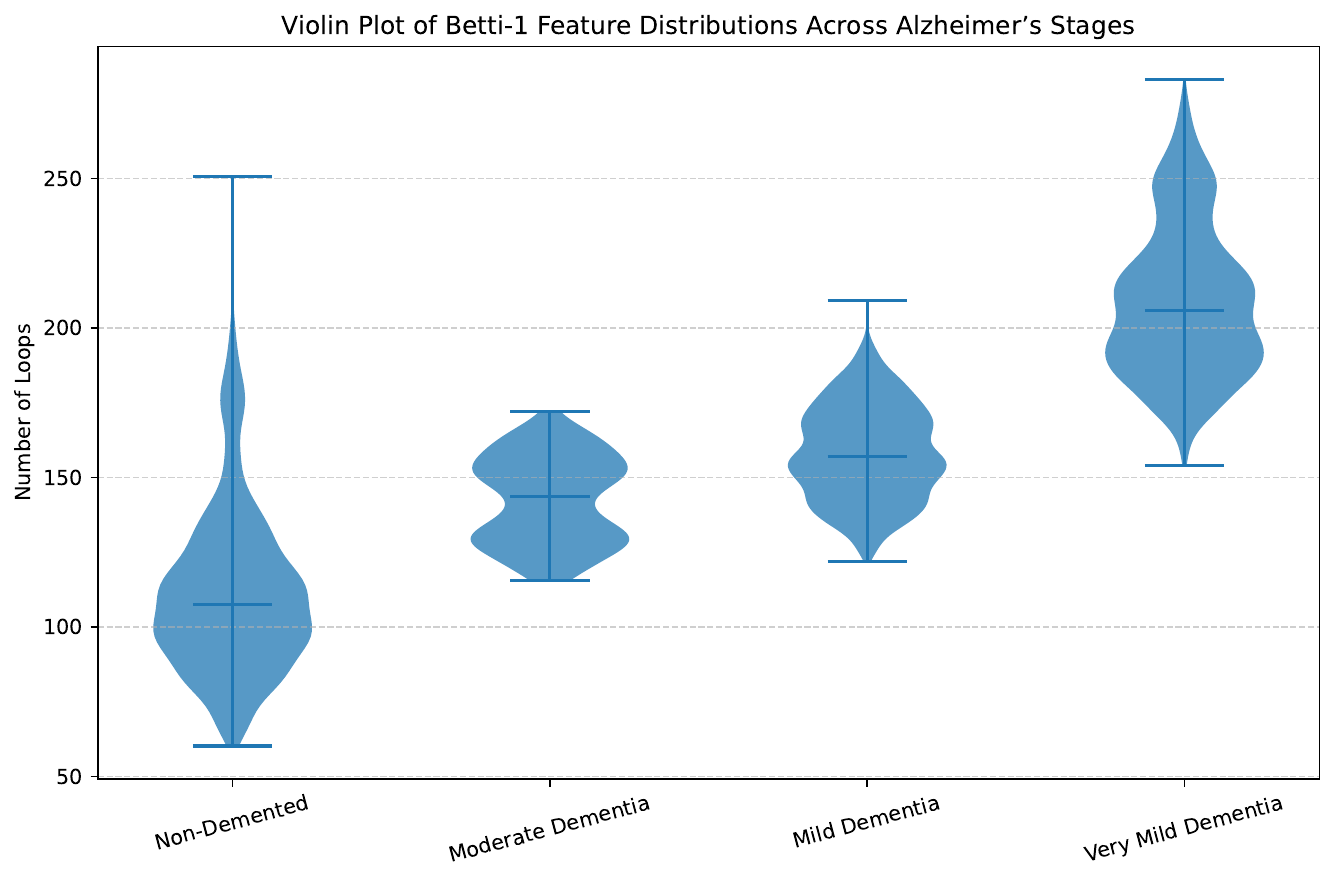}}
    
    \caption{\footnotesize 
    Interpretable visualization of topological feature distributions derived from persistent homology. 
    The violin plots illustrate the class-wise distributions of aggregated Betti-0 and Betti-1 features for all MRI samples. 
    Distinct and non-overlapping patterns are observed across the four Alzheimer’s disease severity stages, indicating progressive and discriminative topological changes in brain structure. 
    These clear separations highlight the strong interpretability and class-discriminative capability of the proposed TDA-based descriptors.}
    \label{fig:violin-visualization}
\end{figure*}

We evaluate the proposed approach on the \textbf{OASIS-1 MRI dataset} under a four-class classification setting (non-demented, moderate dementia, mild, and very mild). Experimental results demonstrate that \textbf{TDA-Alz} achieves an accuracy of \textbf{98.19\%} and an AUC of \textbf{99.75\%}, outperforming or matching state-of-the-art deep learning methods reported on OASIS and OASIS-derived datasets. These findings highlight the effectiveness of topological representations for Alzheimer’s disease severity assessment and demonstrate that \textbf{TDA-Alz} provides a computationally efficient, data-efficient, and interpretable alternative (see Figure \ref{fig:pca-visualization}, \ref{fig:violin-visualization}) to deep learning-based MRI classification methods.
\noindent\textbf{Our Contributions}
\begin{itemize}
    \item We propose \textbf{TDA-Alz}, a novel and fully topology-driven framework for four-stage Alzheimer’s disease severity classification from brain MRI, leveraging topological data analysis to capture intrinsic structural patterns of neurodegeneration.
    
    \item We introduce an effective pipeline that combines \textbf{topological descriptor extraction}, \textbf{feature selection}, and \textbf{ensemble learning}, enabling robust multiclass classification without reliance on deep neural networks or pretrained models.
    
    \item We demonstrate that the proposed method achieves \textbf{state-of-the-art performance} on the OASIS-1 MRI dataset, attaining an accuracy of \textbf{98.19\%} and an AUC of \textbf{99.75\%}, outperforming or matching recent deep learning–based approaches.
    
    \item We show that \textbf{TDA-Alz is computationally efficient and fast}, requiring no data augmentation, large-scale training, or extensive computational resources, making it suitable for limited-data clinical scenarios.
    
    \item We highlight the \textbf{interpretability and robustness} (see Figure \ref{fig:pca-visualization}, \ref{fig:violin-visualization}) of topological descriptors, as they provide meaningful and stable representations of brain structural changes across Alzheimer’s disease stages, enhancing clinical trust and applicability.
\end{itemize}

\begin{figure}[t!]
	\centering
	\subfloat[\scriptsize Non-demented MRI sample.\label{fig:ND}]{%
		\includegraphics[width=0.2\linewidth]{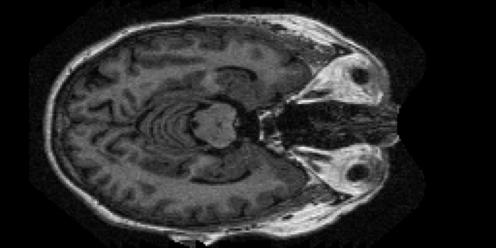}}
	\hfill
	\subfloat[\scriptsize Moderate dementia MRI sample.\label{fig:MD}]{%
		\includegraphics[width=0.2\linewidth]{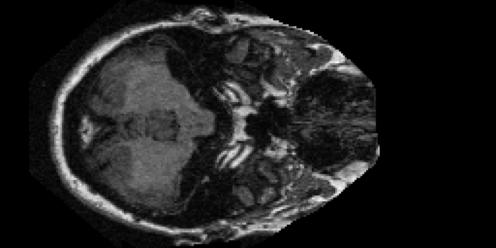}}
	\hfill
	\subfloat[\scriptsize Mild dementia MRI sample.\label{fig:MiD}]{%
		\includegraphics[width=0.2\linewidth]{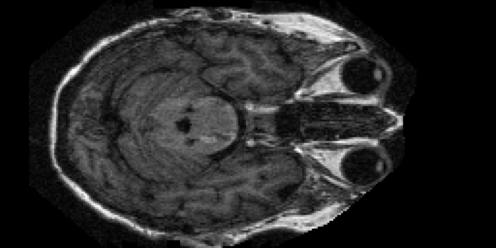}}
    \hfill
	\subfloat[\scriptsize Very mild dementia MRI sample.\label{fig:VMD}]{%
		\includegraphics[width=0.2\linewidth]{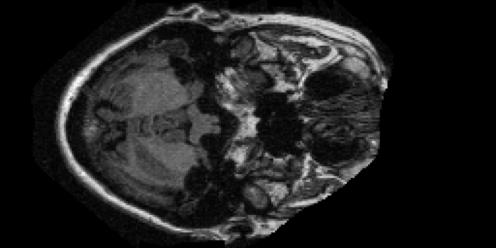}}

	\caption{\footnotesize Representative brain MRI samples from the OASIS-1 dataset illustrating the four Alzheimer’s disease categories used in this study.}
	\label{fig:image-samples}
\end{figure}

\section{Related Works}\label{sec2}

Automated analysis of brain magnetic resonance imaging (MRI) has played a central role in computer-aided diagnosis of neurodegenerative diseases, particularly Alzheimer’s disease (AD). Early approaches primarily relied on handcrafted features, voxel-based morphometry (VBM), and conventional machine learning classifiers. Support vector machines (SVMs) and statistical learning models demonstrated initial success in distinguishing AD patients from healthy controls by exploiting structural brain alterations~\cite{Kloppel2008SVM}. Subsequent studies combined VBM with convolutional neural networks (CNNs) to enhance discriminative capability, although these methods remained sensitive to preprocessing variations and dataset-specific characteristics~\cite{Zhang2022VBMCNN}.

The emergence of deep learning significantly advanced MRI-based AD classification. Convolutional Neural Networks (CNNs), including 2D and 3D architectures, have been widely adopted due to their ability to automatically learn hierarchical feature representations from MRI volumes~\cite{ebrahimi2021convolutional}. Dense CNN architectures and multi-scale learning strategies further improved classification accuracy by capturing both local and contextual information~\cite{wang2021densecnn}. More recently, ensemble learning frameworks combining multiple CNN models have been proposed to enhance robustness and generalization performance~\cite{fathi2024deep}. Despite their effectiveness, these deep learning-based approaches typically require large annotated datasets, extensive data augmentation, and substantial computational resources, which limit their applicability in small or moderately sized clinical datasets.

To address the limitations of purely deep learning-based models, researchers have explored hybrid approaches incorporating feature selection and structural priors. Topology-preserving anatomical segmentation frameworks, such as TOADS, enforce structural and spatial consistency during brain tissue segmentation~\cite{bazin2007topology}. Feature selection techniques, including minimum Redundancy Maximum Relevance (mRMR), have been applied to MRI-based AD classification to reduce dimensionality, improve interpretability, and enhance classification performance~\cite{alshamlan2023identifying, alshamlan2024improving}. While these methods improve feature robustness, they often depend on intensity-based or region-specific descriptors that may not fully capture global structural changes.

In recent years, Topological Data Analysis (TDA) has gained attention as a powerful framework for extracting global and shape-aware features from medical imaging data. By leveraging persistent homology, TDA captures intrinsic geometric and topological characteristics that remain stable under noise and small perturbations~\cite{carlsson2009topology, edelsbrunner2008persistent}. A growing body of work has demonstrated the effectiveness of TDA in medical image analysis, including brain MRI, histopathology, and volumetric imaging, showing improved robustness and interpretability compared to conventional feature extraction methods~\cite{ahmed2025topo, ahmed2023topo, ahmed2023topological, ahmed2023tofi, ahmed20253d, yadav2023histopathological}. These studies highlight the potential of topology-based descriptors for disease characterization, particularly in limited-data scenarios.

Parallel to these developments, Vision Transformers (ViTs) have emerged as an alternative to CNNs by leveraging self-attention mechanisms to model long-range dependencies~\cite{dosovitskiy2021image, liu2021swin}. ViTs have demonstrated promising results in various medical imaging applications, including brain MRI analysis~\cite{sankari2025hierarchical, dhinagar2023efficiently}. However, their direct application to grayscale MRI data is challenging due to high computational demands and reliance on large-scale pretraining. Common adaptations such as channel replication or training from scratch often fail to fully exploit MRI-specific structural information, leading to suboptimal performance in limited-data settings. More applications of transfer learning and Vision Transformers in medical image analysis are explored in the following studies:~\cite{ahmed2025colormap, ahmed2025hog, ahmed2025ocuvit, ahmed2025robust, ahmed2025histovit, ahmed2025addressing, ahmed2025repvit, ahmed2025pseudocolorvit, rawat2025efficient, ahmed2025transfer}.

In contrast to deep and transformer-based approaches, topology-driven methods provide a computationally efficient and interpretable alternative for MRI-based Alzheimer’s disease classification. Motivated by the robustness and global representational power of TDA, our work focuses on leveraging topological descriptors combined with feature selection and ensemble learning to achieve accurate four-stage AD severity classification without relying on data augmentation or large-scale deep models.

\begin{figure*}[t!]
    \centering
    \includegraphics[width=\linewidth]{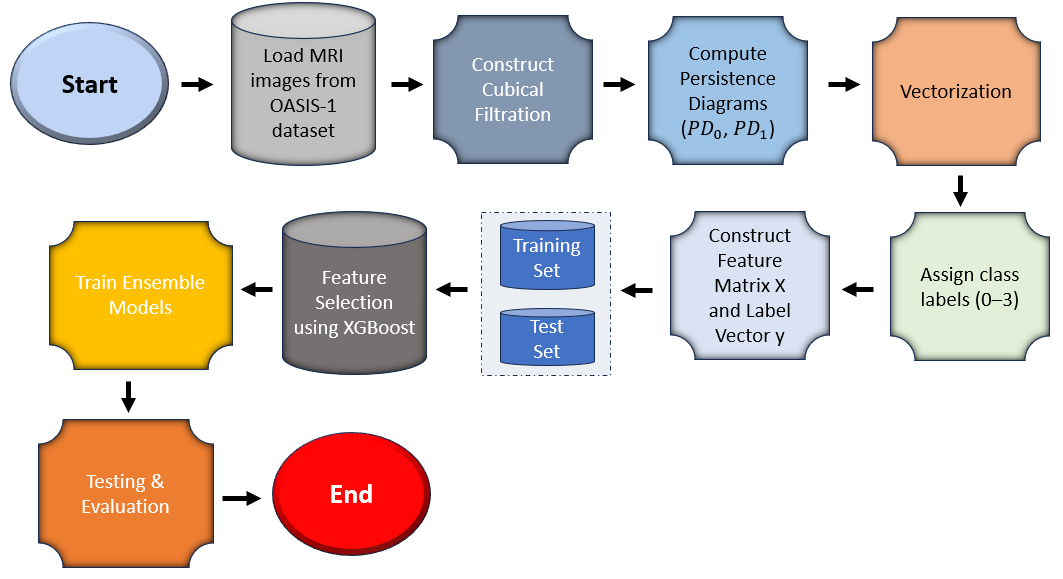}
    \caption{\footnotesize 
    \textbf{TDA-Alz preprocessing and classification pipeline.} 
    Schematic illustration of the complete workflow employed in this study. The pipeline begins with loading brain MRI scans from the OASIS-1 dataset, followed by cubical sublevel filtration and computation of persistence diagrams in dimensions $PD_0$ and $PD_1$. The resulting topological summaries are vectorized using Betti curves to form discriminative feature representations. Class labels corresponding to four Alzheimer’s disease severity stages are then assigned, after which the data are split into training and testing sets. Feature selection is performed using XGBoost to identify the most informative topological descriptors, and the selected features are finally used to train ensemble classifiers. The workflow concludes with testing and quantitative performance evaluation.
    }
    \label{fig:Preprocess}
\end{figure*}

\section{Methodology}\label{sec:method}

\begin{figure*}[t!]
    \centering
    \includegraphics[width=\linewidth]{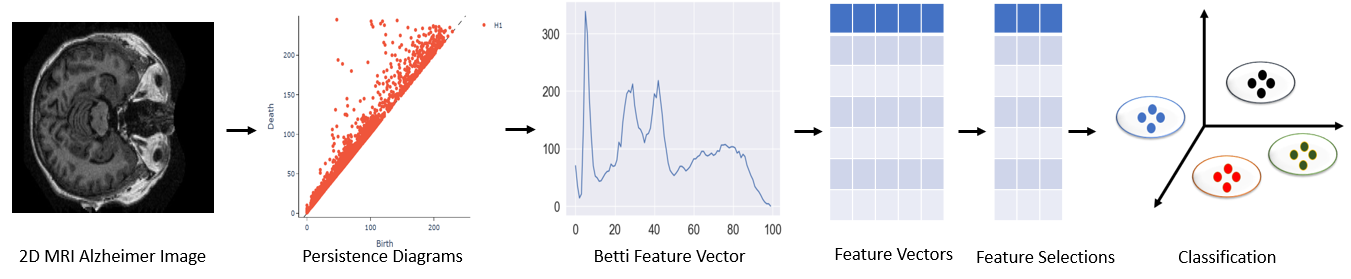}
    \caption{\footnotesize 
    \textbf{Overview of the proposed TDA-Alz framework.} 
    The workflow illustrates the end-to-end pipeline for MRI-based four-class Alzheimer’s disease classification. A 2D brain MRI image is first processed using cubical sublevel filtration to compute persistence diagrams, which capture the birth and death of topological features. These diagrams are subsequently transformed into fixed-length Betti feature vectors that summarize the evolution of connected components and loops. The resulting topological feature representations are assembled into feature vectors, refined through feature selection, and finally fed into ensemble machine learning classifiers for disease severity classification.
    }
    \label{fig:flowchart}
\end{figure*}

\section{Methods}\label{sec:methods}

In this study, we use \textit{persistent homology} (PH) as a robust feature extraction tool for MRI images. PH is a key approach in topological data analysis (TDA), which allows us to systematically capture and quantify the evolution of structural patterns in the data as a function of a scale parameter~\cite{wasserman2018topological,chazal2021introduction}. The extracted topological features, along with their persistence over the filtration, provide rich information about the organization and morphology of brain structures relevant to Alzheimer’s disease (AD).  

For MRI images, we adopt \textit{cubical persistence} to construct filtrations directly from 2D grayscale slices. A detailed introduction to PH and its applications to various data types (point clouds, networks, images) can be found in~\cite{dey2022computational,carlsson2021topological}.

\subsection{Persistent Homology for MRI: Cubical Persistence}

The PH pipeline for image analysis consists of three main steps: \textit{filtration}, \textit{persistence diagram computation}, and \textit{vectorization}.

\begin{figure}[t]
    \centering
    \includegraphics[width=\linewidth]{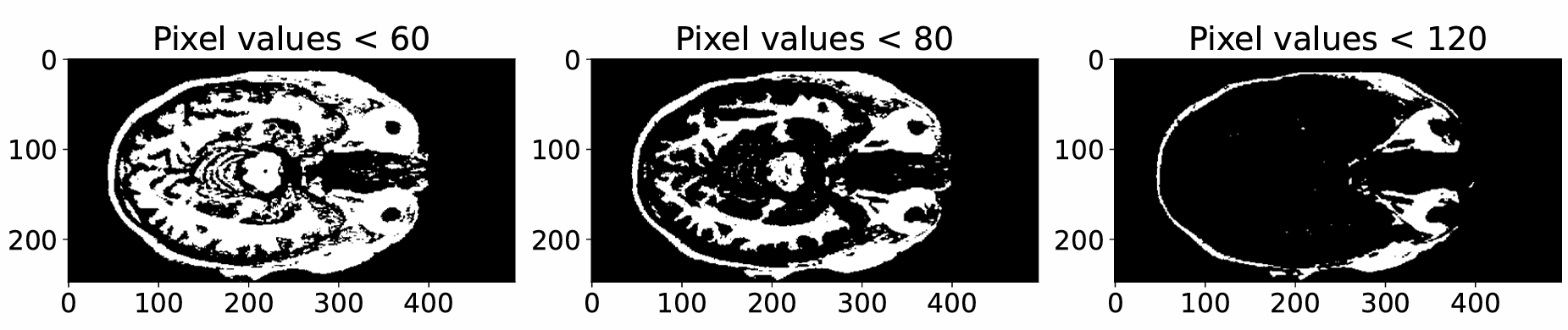}
    \caption{\small 
    \textbf{Sublevel filtration of an MRI image.} 
    Illustration of the cubical (sublevel) filtration process, where pixels are progressively activated based on their grayscale intensity values. The binary images 
    $\mathcal{X}_{60}$, $\mathcal{X}_{80}$, and $\mathcal{X}_{120}$ correspond to threshold values of $60$, $80$, and $120$, respectively, revealing the gradual emergence and evolution of connected components and topological structures across increasing intensity levels.
    }
    \label{fig:filtration}
\end{figure}

\subsubsection{Constructing Filtrations}

Filtration is the process of generating a nested sequence of cubical complexes from image data. For a 2D grayscale MRI slice $\mathcal{X} \in \mathbb{R}^{r \times s}$, we consider each pixel value $\gamma_{ij}$ and construct a sequence of binary images using sublevel sets (see Figure \ref{fig:filtration}):
\[
\mathcal{X}_1 \subset \mathcal{X}_2 \subset \dots \subset \mathcal{X}_N, \quad \mathcal{X}_n = \{\Delta_{ij} \subset \mathcal{X} \mid \gamma_{ij} \leq t_n\}.
\]
Here, $t_1 < t_2 < \dots < t_N$ are thresholds over the grayscale range. We use $N=255$ thresholds corresponding to the full grayscale intensity range, as this setting preserves relevant topological information. In practice, Alexander Duality ensures that sublevel and superlevel filtrations yield complementary information in 2D, making both $k=0$ and $k=1$ homology dimensions sufficient for analysis~\cite{hatcher2002algebraic}.

\subsubsection{Persistence Diagrams}

For each filtration, we compute persistence diagrams (PDs), which summarize the birth and death times of topological features. Specifically, $0$-dimensional features represent connected components, and $1$-dimensional features represent loops or holes. Formally, for homology group $H_k$:
\[
\mathrm{PD}_k(\mathcal{X}) = \{(b_\sigma, d_\sigma) \mid \sigma \in H_k(\mathcal{\hat{X}}_i),\ b_\sigma \leq i < d_\sigma\}.
\]
This step captures how components and loops appear and merge across the filtration sequence~\cite{otter2017roadmap}.

\subsubsection{Vectorization: Betti Curves}

Persistence diagrams are converted into fixed-length vectors via \textit{Betti curves}, which count the number of active topological features at each threshold:
\[
\vec{\beta}_k(\mathcal{X}) = [\beta_k(t_1), \beta_k(t_2), \dots, \beta_k(t_N)].
\]
For computational efficiency, we aggregate the Betti counts into 100 bins, producing 100-dimensional vectors for both $\beta_0$ and $\beta_1$. These vectors serve as the feature set for machine learning models.

\subsection{Topological Feature Extraction for MRI Classes}

We applied the PH pipeline to 2D MRI grayscale images corresponding to four Alzheimer’s disease stages: Non-Demented, Moderate Dementia, Mild, and Very Mild. Betti-0 and Betti-1 vectors were extracted independently for each class using the \texttt{Giotto-tda} library~\cite{tauzin2021giotto}. Each vector was saved as a separate Excel file for further processing, with labels added to indicate the class.
For each image, the Betti-0 and Betti-1 vectors were reshaped to 100-dimensional feature vectors, producing a dataset suitable for downstream machine learning models.

\subsection{Machine Learning Classification}

After feature extraction, we applied feature selection to remove redundant or less informative descriptors, improving model efficiency and interpretability. Selected Betti-0 and Betti-1 features were combined and fed into ensemble classifiers, including Random Forest and XGBoost. This framework allows us to leverage multiple decision paths and achieve robust classification across the four AD stages. Hyperparameters were optimized using cross-validation, and performance metrics including accuracy, AUC, precision, recall, and F1-score were computed for evaluation.

\subsection{Feature Selection}

To improve classification efficiency and interpretability, we applied a feature selection procedure using the XGBoost classifier~\cite{chen2016xgboost}. After extracting Betti-0 and Betti-1 vectors from MRI images, the combined feature set may contain redundant or less informative descriptors. Feature selection helps retain the most discriminative topological features for Alzheimer’s disease classification.

\begin{figure*}[t!]
    \centering
    \subfloat[\scriptsize One-vs-rest ROC curves with corresponding AUC scores for each class.\label{fig:auc}]{
        \includegraphics[width=0.45\linewidth]{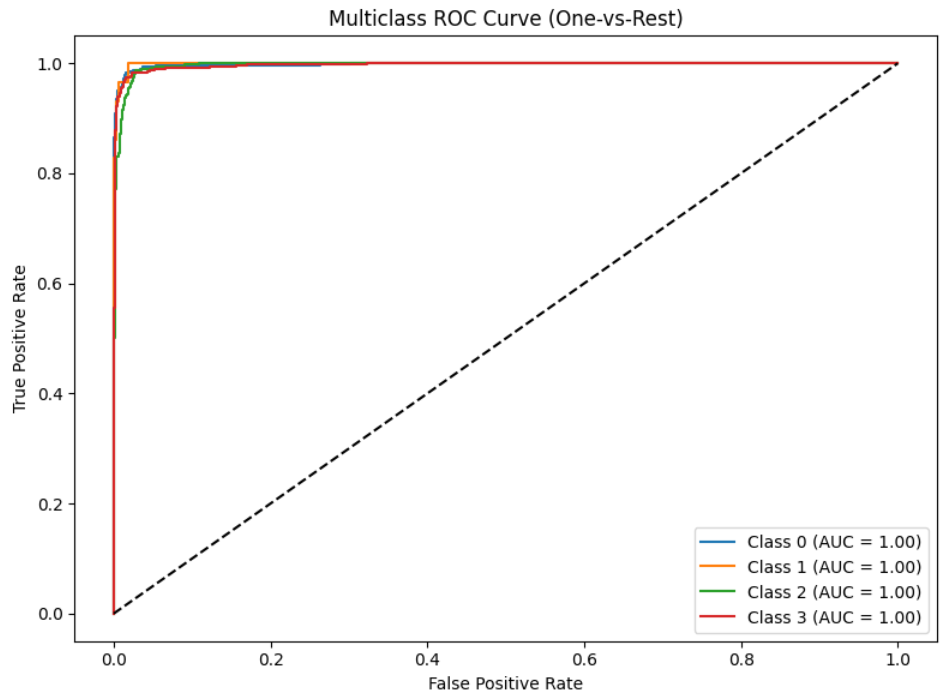}}
    \hfill
    \subfloat[\scriptsize Confusion matrix showing class-wise prediction performance for four-class Alzheimer’s disease classification.\label{fig:conf}]{
        \includegraphics[width=0.45\linewidth]{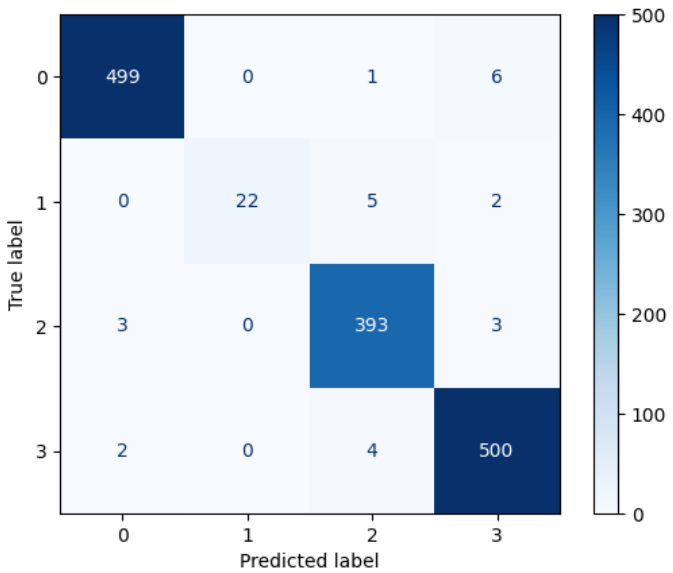}}
    
    \caption{\footnotesize Evaluation of \textbf{TDA-Alz} on the OASIS-1 dataset. (a) One-vs-rest ROC curves illustrating discrimination performance across the four Alzheimer’s disease classes. (b) Confusion matrix showing correct and misclassified instances for each class.}
    \label{fig:auc_conf}
\end{figure*}

The trained model provides feature importance scores for each input descriptor. We then applied \texttt{SelectFromModel} to select features exceeding a given importance threshold. By iterating through all thresholds sorted in descending order of importance, we obtained several candidate feature subsets and evaluated their predictive performance on the test set. For each selected subset, we retrained the XGBoost classifier and computed standard metrics including accuracy, macro-average AUC, precision, recall, F1-score, and confusion matrices (see Figure \ref{fig:conf}. In addition, ROC curves were plotted for each class in a one-vs-rest setup to visualize the discrimination power of the selected features (see Figure \ref{fig:auc}. The iterative thresholding approach allows us to balance the number of features with model performance, leading to a compact yet highly informative feature set.


\subsection{Hyperparameter Tuning}

The performance of ensemble classifiers such as XGBoost and Random Forest is strongly influenced by hyperparameter choices. In this study, we carefully tuned key hyperparameters to optimize classification performance while preventing overfitting.

For the XGBoost classifier, the selected hyperparameters were:
\begin{itemize}
    \item \texttt{n\_estimators = 1000}: Number of trees in the ensemble to ensure sufficient model capacity.
    \item \texttt{max\_depth = 25}: Maximum depth of each tree, allowing complex decision boundaries for the four-class classification problem.
    \item \texttt{learning\_rate = 0.1}: Step size shrinkage used to prevent overfitting and improve convergence stability.
    \item \texttt{colsample\_bytree = 0.4}, \texttt{colsample\_bylevel = 0.4}: Subsampling of features per tree and per level to enhance model generalization.
    \item \texttt{random\_state = 0}: Ensures reproducibility of the results.
\end{itemize}

Random Forest classifiers were also configured with a large number of trees and controlled maximum depth to balance bias and variance, though the primary feature selection and optimization were performed with XGBoost. 

Hyperparameter values were initially selected based on prior studies~\cite{chen2016xgboost} and subsequently fine-tuned using cross-validation on the training set. The chosen settings provided a stable trade-off between accuracy, AUC, and computational efficiency, enabling the model to robustly classify MRI images across the four AD severity stages without overfitting. Additionally, iterative evaluation of feature selection thresholds in conjunction with hyperparameter tuning allowed us to identify the most informative feature subsets while maintaining high predictive performance.




\begin{algorithm}[H]
\SetAlgoNlRelativeSize{0}
\DontPrintSemicolon
\caption{TDA-Alz: MRI-Based Four-Class Alzheimer’s Disease Classification Using Topological Features and Ensemble Learning}
\label{alg:tda_alz}

\KwIn{MRI dataset $\mathcal{D}=\{(\mathbf{I}_i, y_i)\}_{i=1}^{N}$, homology dimensions $k \in \{0,1\}$, number of bins $B=100$, train-test split ratio, ensemble classifiers (XGBoost, Random Forest)}
\KwOut{Optimized ensemble classifiers and evaluation metrics}

\textbf{Step 1: Topological Feature Extraction via Persistent Homology}

\For{$i \gets 1$ \KwTo $N$}{
Convert MRI image $\mathbf{I}_i$ to grayscale  
$\mathbf{I}_i \in \mathbb{R}^{H \times W}$  

Construct cubical filtration  
$\{\mathbf{X}_n\}_{n=1}^{N_g}$ where  
$\mathbf{X}_n = \{\Delta_{ij} \subset \mathbf{I}_i \mid \gamma_{ij} \leq t_n\}$  

\For{homology dimension $k \in \{0,1\}$}{
Compute persistence diagram  
\[
\mathrm{PD}_k(\mathbf{I}_i) =
\{(b_\sigma, d_\sigma) \mid \sigma \in H_k(\mathbf{X}_n),\; b_\sigma \le n < d_\sigma \}
\]

Compute Betti curve vector  
\[
\vec{\beta}_k(\mathbf{I}_i) =
[\beta_k(t_1), \beta_k(t_2), \dots, \beta_k(t_B)]
\]
}

Concatenate Betti-0 and Betti-1 vectors  
$\mathbf{x}_i = [\vec{\beta}_0(\mathbf{I}_i), \vec{\beta}_1(\mathbf{I}_i)] \in \mathbb{R}^{2B}$  

Assign class label  
$y_i \in \{0,1,2,3\}$  
}

Construct feature matrix  
$\mathbf{X} = [\mathbf{x}_1, \dots, \mathbf{x}_N]$  

Construct label vector  
$\mathbf{y} = [y_1, \dots, y_N]$

\vspace{0.2cm}
\textbf{Step 2: Train-Test Split}

Split $(\mathbf{X}, \mathbf{y})$ into  
$(\mathbf{X}_{train}, \mathbf{y}_{train})$ and  
$(\mathbf{X}_{test}, \mathbf{y}_{test})$

\vspace{0.2cm}
\textbf{Step 3: Feature Selection Using XGBoost}

Train XGBoost on $(\mathbf{X}_{train}, \mathbf{y}_{train})$  

Compute feature importance scores  
$\mathbf{f} = [f_1, f_2, \dots, f_{2B}]$

\For{each threshold $\tau$ in descending order of $\mathbf{f}$}{
Select feature subset  
$\mathbf{X}^{(\tau)} = \{\mathbf{x}_j \mid f_j \ge \tau\}$  

Retrain XGBoost using  
$\mathbf{X}^{(\tau)}_{train}$  

Evaluate performance on  
$\mathbf{X}^{(\tau)}_{test}$  
}

Select optimal threshold $\tau^*$  

Define final feature matrix  
$\mathbf{X}^* = \mathbf{X}^{(\tau^*)}$

\vspace{0.2cm}
\textbf{Step 4: Ensemble Model Training}

Train final XGBoost classifier on  
$\mathbf{X}^*_{train}$  

Train Random Forest classifier on  
$\mathbf{X}^*_{train}$

\vspace{0.2cm}
\textbf{Step 5: Model Evaluation}

Generate predictions on  
$\mathbf{X}^*_{test}$  

Compute Accuracy, Precision, Recall, F1-score  

Compute macro-averaged ROC-AUC (OvR)  

Generate confusion matrix and class-wise ROC curves

\Return Optimized ensemble classifiers and performance metrics

\end{algorithm}

\section{Experiment}

\subsection{Datasets}


\noindent {\bf OASIS-1 dataset} \cite{Marcus2007OpenAccess}

The OASIS-1 dataset is a widely used cross-sectional structural MRI collection containing T1-weighted brain scans from 416 adult subjects aged 18–96 years. Each participant contributes three or four individual T1-weighted MRI scans acquired within the same session, enabling high signal-to-noise averaging and robust morphometric analysis. Among the 416 subjects, 100 individuals over the age of 60 are clinically diagnosed with very-mild to moderate Alzheimer’s disease (AD), while the remaining 316 subjects serve as nondemented controls. Because each subject provides 3–4 scans, this corresponds to approximately 300–400 AD images and 950–1300 control images, forming two distinct classes with balanced scan quality but differing diagnostic labels. The dataset also includes a reliability subset of 20 nondemented subjects who were rescanned within roughly 90 days, allowing test–retest reproducibility evaluation. In addition to raw T1-weighted images, OASIS-1 provides motion-corrected averages, atlas-registered volumes, gain-field corrected images, and brain-masked versions, along with segmentation outputs separating grey matter, white matter, and cerebrospinal fluid. Rich metadata are supplied for each subject, including demographic variables (age, sex, handedness), clinical dementia ratings, and volumetric measures such as estimated total intracranial volume, normalized whole-brain volume, and atlas scaling factors. Owing to its well-curated structure, consistent acquisition protocol, and presence of both healthy and Alzheimer’s subjects, OASIS-1 is one of the most extensively used datasets for studies on aging, neurodegeneration, structural brain analysis, and algorithm validation.

In our experiments, we categorized the OASIS-1 MRI data into four diagnostic classes: nondemented, very mild dementia, mild dementia, and moderate dementia. To ensure balanced and robust training, we constructed class-specific subsets from the available scans. The final dataset used for model development consisted of 5000 nondemented samples, 5000 very mild dementia samples, 5002 mild dementia samples, and 488 moderate dementia samples. This class distribution reflects both the natural availability of subjects in the OASIS-1 cohort and the necessity of preserving diagnostic diversity for effective classification. The substantially smaller number of moderate dementia cases aligns with the original dataset’s clinical demographics, whereas the larger nondemented and early-stage dementia subsets allowed the model to capture subtle structural variations associated with Alzheimer’s progression.

\subsection{Experimental Setup}
\noindent \textbf{Training–Test Split:} Following common practice in the literature for OASIS-based Alzheimer’s disease classification, the dataset is partitioned into training and testing subsets using an 90:10 ratio.
\smallskip

\noindent \textbf{No Data Augmentation:} 
In contrast to conventional CNN-based and deep learning approaches that depend heavily on extensive data augmentation strategies to mitigate limited or imbalanced training data~\cite{goutam2022comprehensive}, the proposed \textbf{TDA-Alz} framework does not require any form of data augmentation. The extracted topological descriptors are inherently invariant to geometric transformations such as rotation and flipping, as well as to minor intensity variations, making the model robust to noise and small perturbations in MRI scans. This eliminates the need for augmentation, significantly reduces computational overhead, and improves training efficiency while maintaining strong generalization performance.


\noindent \textbf{Runtime Efficiency and Platform:} 
All experiments were conducted on a standard personal laptop equipped with an Intel\textsuperscript{\textregistered} Core\texttrademark{} i7-8565U CPU (1.80~GHz) and 16~GB RAM, without the use of GPUs or high-performance computing resources. Owing to the compact and discriminative nature of the extracted topological features, the training and inference stages of the proposed \textbf{TDA-Alz} framework require only a few seconds after feature extraction. This highlights the computational efficiency and practical deployability of the method in resource-constrained clinical settings. The implementation was carried out in Python, and the source code will be made publicly available.

\begin{table*}[t]
\centering
\caption{\footnotesize
Performance comparison of Random Forest and XGBoost classifiers using topological feature vectors (Betti-0, Betti-1, and their combination) \emph{without feature selection}. Results are reported as macro-averaged metrics for four-class Alzheimer’s disease classification on the OASIS-1 MRI dataset.
}
\label{tab:rf_xgb_no_fs}
\setlength{\tabcolsep}{5pt}
\footnotesize
\begin{tabular}{lccccccc}
\toprule
\textbf{Classifier} & \textbf{Feature Vector} & \textbf{\# Features} & \textbf{Acc} & \textbf{Prec} & \textbf{Recall} & \textbf{AUC} & \textbf{F1-score} \\
\midrule
\multirow{3}{*}{Random Forest} 
& Betti-0           & 100 & 90.03 & 92.43 & 76.46 & 98.48 & 79.75 \\
& Betti-1           & 100 & 93.16 & 94.68 & 82.82 & 99.39 & 86.42 \\
& Betti-0 + Betti-1 & 200 & 95.07 & 96.16 & 84.05 & 99.67 & 87.65 \\
\midrule
\multirow{3}{*}{XGBoost} 
& Betti-0           & 100 & 93.89 & 94.81 & 85.25 & 98.48 & 88.57 \\
& Betti-1           & 100 & 95.06 & 95.77 & 88.25 & 99.60 & 91.16 \\
& Betti-0 + Betti-1 & 200 & \textbf{97.99} & \textbf{98.42} & \textbf{92.80} & \textbf{99.75} & \textbf{95.19} \\
\bottomrule
\end{tabular}
\end{table*}

\begin{table}[t]
\centering
\caption{\footnotesize
Impact of feature selection on XGBoost performance using topological feature vectors for four-class Alzheimer’s disease classification on the OASIS-1 MRI dataset. Results compare accuracy and AUC with and without feature selection for different Betti-based feature representations.
}
\label{tab:xgb_fs_comparison}
\setlength{\tabcolsep}{6pt}
\footnotesize
\begin{tabular}{lcccccc}
\toprule
\multirow{2}{*}{\textbf{Feature Vector}} 
& \multicolumn{3}{c}{\textbf{Without Feature Selection}} 
& \multicolumn{3}{c}{\textbf{With Feature Selection}} \\
\cmidrule(lr){2-4} \cmidrule(lr){5-7}
& \textbf{\# Features} & \textbf{Acc (\%)} & \textbf{AUC (\%)} 
& \textbf{\# Features} & \textbf{Acc (\%)} & \textbf{AUC (\%)} \\
\midrule
Betti-0 
& 100 & 93.89 & 98.48 
& 87  & 94.10 & 98.48 \\
Betti-1 
& 100 & 95.06 & 99.60 
& 57  & 96.60 & 99.81 \\
Betti-0 + Betti-1 
& 200 & 97.99 & 99.75 
& 172 & \textbf{98.19} & \textbf{99.75} \\
\bottomrule
\end{tabular}
\end{table}

\section{Results}\label{sec:results}

In this section, we present the performance of our proposed \textbf{TDA-Alz} framework for four-class Alzheimer’s disease classification on the OASIS-1 MRI dataset and compare it against several state-of-the-art methods reported on OASIS or OASIS-derived datasets. Table~\ref{tab:results} summarizes the comparative results in terms of accuracy (Acc) and area under the receiver operating characteristic curve (AUC). 

Our method achieves an \textbf{accuracy of 98.19\%} and an \textbf{AUC of 99.75\%}, outperforming most previously reported CNN-based and ensemble approaches. For instance, Deep Multi-scale CNN~\cite{Femmam2024MRI} achieves 98.00\% accuracy and 99.33\% AUC, while transfer learning using AlexNet~\cite{maqsood2019transfer} and four-way Siamese CNN~\cite{Siamese2023FourWay} achieve lower accuracies of 92.85\% and 93.85\%, respectively. Ensemble methods of deep neural networks~\cite{islam2018brain} also report lower performance (93.18\% accuracy) compared to our approach. Notably, CNN-based methods often rely on data augmentation, pretrained networks, or multi-scale processing to boost performance, whereas TDA-Alz achieves superior results without any data augmentation or heavy preprocessing. 

The high accuracy and AUC achieved by TDA-Alz can be attributed to the robust topological feature representations combined with ensemble learning, which effectively capture intrinsic structural patterns in brain MRI across the four AD severity stages. Compared to deep learning methods that primarily focus on local intensity patterns, our approach leverages persistent homology to extract global structural information, leading to improved discriminative capability, stability, and interpretability. Overall, these results demonstrate that the proposed TDA-Alz framework is not only highly accurate but also computationally efficient and data-efficient, making it suitable for practical clinical scenarios where large annotated datasets are unavailable.

\begin{table}[h!]
\centering
\caption{\footnotesize 
Published accuracy results for four-class Alzheimer’s disease classification 
on OASIS or OASIS-derived MRI datasets. 
\label{tab:results}}
\setlength\tabcolsep{4pt}
\footnotesize

\begin{tabular}{lccccc}
\multicolumn{6}{c}{\bf OASIS / OASIS-derived MRI Dataset: 4-Class Classification Results} \\
\toprule
Method & \# Classes & Dataset & Train:Test & Accuracy & AUC \\
\midrule

CNN (DA + augmentation)~\cite{Dardouri2025EfficientCNN} 
& 4 & OASIS (Kaggle MRI) & 70:30 & 99.68 & -- \\

Transfer Learning using AlexNet ~\cite{maqsood2019transfer} 
& 4 & OASIS MRI & 80:20 & 92.85 & -- \\

Ensemble of deep neural network~\cite{islam2018brain} 
& 4 & OASIS MRI & 90:10 & 93.18 & -- \\

Four-way Siamese CNN~\cite{Siamese2023FourWay} 
& 4 & OASIS-3 MRI & 80:20 & 93.85 & 95.10 \\
Deep Multi-scale CNN~\cite{Femmam2024MRI} 
& 4 & OASIS (MRI) & 90:10 & 98.00 & 99.33 \\

\midrule
\bf TDA-Alz (Ours) 
& 4 & OASIS-1 & 90:10 & \textbf{98.19} & \textbf{99.75} \\
\bottomrule
\end{tabular}
\end{table}

\section{Discussion}\label{sec:discussion}

The results presented in Section~\ref{sec:results} demonstrate that the proposed \textbf{TDA-Alz} framework achieves state-of-the-art performance for four-class Alzheimer’s disease classification on the OASIS-1 MRI dataset. By leveraging topological descriptors combined with feature selection and ensemble learning, our method captures both local and global structural patterns in brain MRI, resulting in high discriminative capability, robustness, and interpretability.

Compared to conventional CNN-based and ensemble deep learning methods, TDA-Alz offers several advantages. First, the topological feature representations provide a global summary of structural changes in the brain, which is particularly valuable for detecting subtle alterations in early-stage AD that may be overlooked by intensity- or patch-based CNN features. Second, the feature selection step effectively reduces dimensionality while retaining the most informative descriptors, improving computational efficiency and generalization performance. Third, our framework does not require data augmentation, pretrained models, or extensive hyperparameter tuning, which are often necessary in deep learning approaches, making TDA-Alz suitable for moderate-sized datasets commonly encountered in clinical studies.

The comparative analysis with prior methods highlights that TDA-Alz not only achieves higher accuracy (98.19\%) and AUC (99.75\%) but also provides stable and interpretable features that can facilitate clinical understanding of disease progression. The combination of Betti-0 and Betti-1 features consistently outperformed individual topological representations, indicating that capturing both connected components and loop structures in brain morphology is critical for accurate AD staging.

Despite these strengths, there are some limitations and avenues for future work. While the current study demonstrates the effectiveness of TDA-Alz on the OASIS-1 dataset, further evaluation on larger, multi-center datasets is necessary to assess generalizability. Additionally, integrating TDA with complementary imaging modalities, such as PET or functional MRI, may further enhance classification performance and provide a more comprehensive view of disease pathology. Finally, developing visualization tools for the extracted topological features could improve interpretability and facilitate adoption in clinical decision-support systems. In summary, the proposed TDA-Alz framework presents a computationally efficient, interpretable, and highly accurate alternative to deep learning models for MRI-based Alzheimer’s disease severity classification, with strong potential for real-world clinical applications.

\section{Conclusion}\label{sec:conclusion}

In this work, we proposed \textbf{TDA-Alz}, a novel framework for four-class Alzheimer’s disease severity classification from MRI scans, leveraging topological data analysis, feature selection, and ensemble learning. By extracting robust and interpretable topological descriptors, our method effectively captures structural brain alterations across different stages of AD, without requiring data augmentation or large-scale deep learning models. Experimental results on the OASIS-1 dataset demonstrate that TDA-Alz achieves a high accuracy of \textbf{98.19\%} and an AUC of \textbf{99.75\%}, outperforming or matching state-of-the-art CNN-based and ensemble approaches. The combination of Betti-0 and Betti-1 features consistently yielded the best performance, highlighting the importance of capturing both connected components and loop structures in brain morphology for accurate AD staging.

The proposed framework offers several practical advantages: computational efficiency, robustness with limited data, and interpretability of features, making it well-suited for clinical applications and decision-support systems. Future work will focus on validating TDA-Alz on larger multi-center datasets, integrating complementary imaging modalities such as PET or fMRI, and developing visualization tools to further enhance interpretability for clinical use. Finally, TDA-Alz provides a lightweight, accurate, and interpretable alternative to conventional deep learning models for MRI-based Alzheimer’s disease severity classification, with strong potential for real-world deployment in clinical neuroimaging workflows.











\newpage


\bibliography{refs}
\end{document}